\documentclass{article}
\usepackage{ijcai24}

\usepackage{times}
\usepackage{soul}
\usepackage{url}
\usepackage[hidelinks]{hyperref}
\usepackage[utf8]{inputenc}
\usepackage[small]{caption}
\usepackage{graphicx}
\usepackage{amsmath}
\usepackage{amsthm}
\usepackage{booktabs}
\usepackage{algorithm}
\usepackage{algorithmic}
\usepackage[switch]{lineno}
\usepackage{xcolor}

\title{Personalizing explanations of AI-driven hints to users' cognitive abilities: an empirical evaluation}

\author{
Vedant Bahel$^1$
\and
Harshinee Sriram$^1$\And
Cristina Conati$^1$
\affiliations
$^1$The University of British Columbia\\
\emails
\{bvedant, hsriram, conati\}@cs.ubc.ca
}

\begin{document}

\maketitle

\begin{abstract}
We investigate personalizing the explanations that an Intelligent Tutoring System generates to justify the hints it provides to students to foster their learning. The personalization targets students with low levels of two traits, Need for Cognition and Conscientiousness, and aims  to enhance these students' engagement with the explanations, based on prior findings that these students do not naturally engage with the explanations but they would benefit from them if they do. To evaluate the effectiveness of the personalization, we conducted a  user study where we found that our proposed personalization significantly increases our target users' interaction with the hint explanations, their understanding of the hints and their learning. Hence, this work provides valuable insights into effectively personalizing AI-driven explanations for cognitively demanding tasks such as learning.
\end{abstract}

\section{Introduction}

There has been increasing attention to making AI systems more human-centered by enabling them to explain their actions to their \textit{end-users}. There is substantial research showing that such explanations can be effective (e.g., \cite{kulesza2015principles,heimerl2020unraveling,boggess2023explainable,amitai2023explaining,barria2023adaptive}), although the need and effectiveness of the explanations might depend on application-dependent factors such as task criticality (e.g., \cite{bunt2012explanations} and task-complexity \cite{salimzadeh2023missing}. 

There are also results showing that \textit{user differences} might impact if and how an AI system should explain its actions, where differences range from  user groups (e.g., patient versus clinician \cite{kim2024stakeholder}) to individual user characteristics (UCs) such as personality traits and cognitive capabilities pertaining to individuals rather than groups. \cite{conati2021toward,millecamp2019explain,millecamp2020s,kouki2019personalized,naveed2018argumentation,schaffer2019can}. These works suggest that it is important to investigate personalized Explainable AI (XAI), namely enabling AI systems to personalize explanations of their behaviors to their end-users. 

In this paper, we contribute to this line of work by investigating the personalization of the explanations generated by an Intelligent Tutoring System (ITS) to  justify its pedagogical actions. The personalization targets students with low levels of two UCs: Need for Cognition, which measures one's  inclination for  engaging in cognitively demanding tasks \cite{cacioppo1984efficient}); and  Conscientiousness, a  personality trait measuring dedication to goals and effective task management  \cite{rothmann2003big}. 

To the best of our knowledge, only two other works so far  designed and tested explanations that are personalized to UCs. Millecamp et al. \shortcite{martijn2022knowing} focused on personalizing explanations for users interacting with a  music recommender system, which is arguably a less cognitively demanding task than learning with an ITS. Like in our work, Hostetter et al. \shortcite{hostetter2023xai} personalized the explanations generated by an ITS; however, their explanations consisted of one-shot, relatively simple sentences. The explanations in \cite{millecamp2019explain,millecamp2020s} were also relatively simple combinations of texts and graphics. 

We contribute to these existing works by personalizing explanations provided during the cognitively demanding task of learning from an ITS, like in \cite{hostetter2023xai} but involving explanations that are much more complex as they span several "pages" combining text and graphics, that students can access incrementally as they like. The ITS we leverage in this work is the ACSP (Adaptive CSP) applet, and the explanations are provided by the ACSP to justify the  hints that it generates to  help students master how to solve constraint satisfaction problems (CSPs). 
 
We chose to personalize the ACSP explanations to students with low levels of Need for Cognition (N4C) and Conscientiousness based on the results of previous work \cite{conati2021toward} which  explored the need for personalizing XAI in this context. They found that the ACSP explanations improved students' perception of ACSP hints in terms of trust, helpfulness, and intention to reuse. However, they also found that:

\begin{itemize}
    \item Students with low Conscientiousness learned more from the ACSP applet with explanations than without, while the opposite was true for their high Conscientiousness counterparts.
    \item Students with low N4C showed  reduced attention to explanations compared to their high N4C counterparts.
\end{itemize}

These results suggest that students with low N4C and low Conscientiousness (LNLC from now on) could benefit from personalization that encourages increased interaction with the ACSP explanations to counteract their tendency to disregard them (due to low N4C) and potentially improve their learning (due to low Conscientiousness). In this work, we implement this personalization by presenting the ACSP explanations more proactively and adding prompts encouraging LNLC students to stay with the explanations if they try to dismiss them too soon. We formally evaluate  our proposed personalization with a user study and find that it significantly increases the interaction of LNLC students with the explanations, their perceived understanding of the ACSP hints, and their learning from the ACSP, compared to the original explanations in   \cite{conati2021toward}.

To summarize, this work contributes further evidence on the importance of personalizing explanations of AI behaviors to user differences, looking for the first time at explanations that are interactive, complex, and provided during a cognitively demanding task (i.e., learning)

\section{Related work}

There is increasing evidence highlighting the impact of individual user differences on  XAI effectiveness.  For instance, Kouki et al., \shortcite{kouki2019personalized} performed a crowd-sourced study that revealed that users' preferences for item-centric, user-centric, or socio-centric explanations in a recommender system depend on their levels of the \textit{Neuroticism} personality trait. Naveed et al., \shortcite{naveed2018argumentation} observed that users' decision-making style (rational vs. intuitive) affects their preferences for explanations of recommendations when purchasing a camera (e.g., content-based or item-based). Schaffer et al., \shortcite{schaffer2019can} discovered that explanations of an AI agent's suggestions during a decision game were beneficial exclusively for users with reported low game proficiency. Millecamp et al., \shortcite{millecamp2019explain,millecamp2020s} investigated the impact of several user characteristics on the effectiveness of explanations  for song recommendations. The recommendations were generated based on how well a song fits users' preferences along six song attributes (e.g., popularity and dace-ability), and for each recommendation, the corresponding  explanation  visualized this fit. Their works found that the effectiveness of these explanations depends on users' levels of openness (one's tendency to be open to new ideas and experiences), N4C, and musical sophistication (MS). Conati et al., \shortcite{conati2021toward} (the work that this paper builds upon) uncovered the impact of N4C, Conscientiousness, and Reading Proficiency on the effectiveness of explanations generated to justify the hints generated by the original ACSP applet, as mentioned earlier.

However, our focus is on implementing and assessing the effectiveness of personalized explanations and works in this area are comparatively limited. For instance, Kim et al., \shortcite{kim2024stakeholder} studied personalized explanations of the behavior of an AI-driven system (simulated) that helps patients and their clinicians manage a speech impairment. The personalization involved having two different levels of detail, with the more detailed explanations being designed to suit clinicians. Both groups saw both explanations, and they were asked which one they preferred. Each group preferred the explanations designed for them. 

While this work focused on personalizing explanations to general user groups, Hostetter et al., \shortcite{hostetter2023xai} and
Millecamp et al., \shortcite{martijn2022knowing} go beyond and look at user characteristics (UCs) pertaining to individuals rather than groups.
Hostetter et al., \shortcite{hostetter2023xai} designed and evaluated personalized explanations for the suggestions generated by an ITS that helps students solve probability exercises. The suggestions include solving an exercise independently, in collaboration with the ITS, or having the ITS solve it. The explanations conveyed why it is useful to follow each suggestion, and their content was personalized to the student’s learning attitude. Specifically, for students with a learning-oriented attitude, the explanation emphasized that following the suggestion would help them learn better. for students with an efficiency-oriented attitude, the explanation emphasized the suggestion would help them finish the exercise faster. The explanations were conveyed with simple sentences. The results in \cite{hostetter2023xai} showed that these personalized explanations improve student learning.    

Millecamp et al., \shortcite{martijn2022knowing} built upon prior work in the context of song recommendations \cite{millecamp2019explain,millecamp2020s} by investigating explanations that were personalized to the three user traits that were previously identified as impacting explanation effectiveness in this context i.e., N4C, MS, and openness. The personalization involved both the delivery method (explanations were delivered either upfront with the recommendation or on-demand) and how much detail was included (i.e., just a sentence stating the similarity percentage, an additional bar graph showing the fit for all attributes, or an additional scatterplot). Their results indicated that users with low MS preferred the text-only explanations and those with low openness preferred the explanations that included the bar graph.

The personalization we use in our work is similar to the upfront versus on-demand methods in \cite{martijn2022knowing}, but we investigated it in a very different context, involving both a more complex user task (learning) and much more complex explanations due to the complexity of the underlying AI, described in the following section. 

\begin{figure*}[t]
 \centering
  \includegraphics[width=0.8\textwidth]{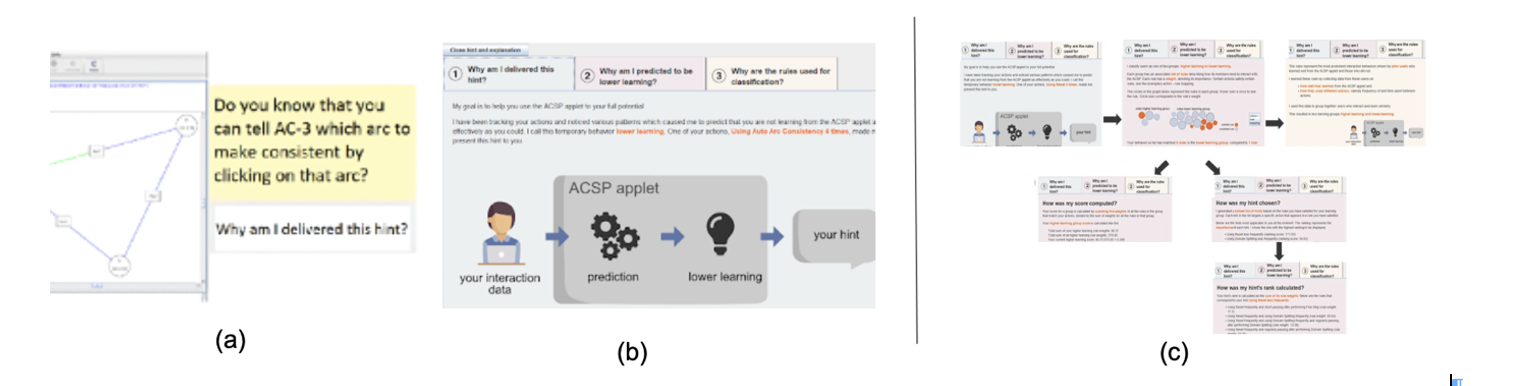}
  \caption{(a) An example of ACSP hint; (b) The first page of the explanation interface that is activated by clicking on “Why I am delivered this hint?” at the bottom of the hint; (c) A schematic view of the six incremental explanations available.}
\end{figure*}

\section{The ACSP applet and its original explanation interface}
The ACSP applet is an ITS that provides tools and AI-driven personalized hints to explore the workings of the AC-3 algorithm to solve constraint satisfaction problems (CSPs). The personalized hints scaffold a more effective usage of the available ACSP tools. For instance, Fig. 1a shows a hint that suggests to the student a specific applet functionality that they have not used so far. The hints are generated leveraging FUMA (Framework for User Modeling and Adaptation) \cite{kardan2015providing,kardan2017data}, a framework for modeling and supporting open-ended exploratory interactions that combine several AI modules such as clustering, association rule mining, and rule-based classification. 

When a student receives a hint, an interface action button at the bottom of it (Fig. 1a) allows them to access explanations of why and how it was generated. Using the interface action brings up the first of a series of incremental explanations (Fig. 1b). These explanations are automatically generated from the ACSP user model, and they are structured around 6 pages (schematically shown in Fig. 1c) that explain different aspects of the underlying AI. The first three pages (\textit{"why"} pages, accessible via the tabs shown in Fig. 1b) justify specific aspects of the rationale for a hint: “Why am I delivered this hint?”, “Why am I predicted to be lower learning?”, and “Why are the rules used for classification?”. From the second why page, the user can access three pages (\textit{"how"} pages, see Fig. 1c) that detail how specific aspects were computed: “How was my hint chosen?”, “How was my score computed”, and “How was my hint’s rank calculated?”. Full details on the original explanation interface can be found in \cite{conati2021toward}.

\begin{figure*}[t]
 \centering
  \includegraphics[width=0.8\textwidth]{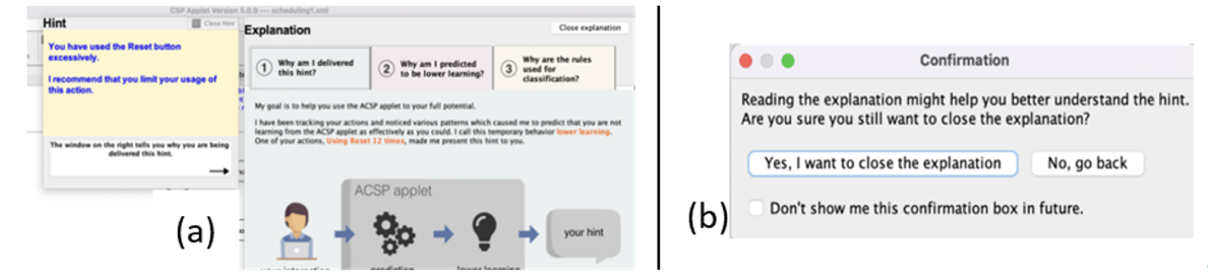}
  \caption{(a) First explanation page appearing upfront with a hint; (b) Confirmation box.}
\end{figure*}

\section{Personalized explanation interface }
As we discussed in the introduction, our goal is to find ways to increase LNLC users' interaction with the ACSP explanation. We do so with modifications to the original explanation interface aimed at increasing (i) access to the interface, and (ii) time spent in it.

\textbf{Increasing access:} We tried two different approaches to stimulate access to the explanations when a hint is provided. The first approach involved disabling the possibility to close the hint for a brief period, during which the only option users have is to open the explanation interface. The second approach was to have the first page of the explanation interface open automatically when a hint is shown (upfront delivery, similar to \cite{martijn2022knowing}), instead of waiting for users to explicitly click on the "why this hint” interface action button. We compared these approaches in a pilot study, which revealed that users favored the upfront delivery approach, which we then adopted moving forward. 
For the upfront delivery approach, users from the pilot study indicated that, when the hint and explanation windows were delivered together, the latter appeared to be more prominent. This led to them looking at the explanations for the hint before looking at the hint, resulting in confusion. To address this issue, we enhanced the visibility of the hint by changing its font size and color, and we added a sentence below the hint saying, “The window on the right tells you why you are being delivered the hint” (Fig. 2a). A follow-up pilot study showed that these minor adjustments helped users first pay attention to the hint and then to its explanations.

\textbf{Increasing time spent in the explanation interface:} To achieve this objective, we added a confirmation box that appears when a user tries to close the explanation interface too early after accessing it. To determine how early is “too early”, we used the median time High N4C users spent in the explanation interface in the previous study \cite{conati2021toward} (i.e., 27 seconds) as they were the ones who paid substantial attention to the explanations. The text in the confirmation box conveys the value of reading the explanations and then asks the user if they would still like to close it (Fig. 2b).

We ran a user study to evaluate whether these changes to the original ACSP explanation interface manage to increase LNLC users' interaction with it and subsequently their learning. 

\section{User study}
The user study had the same between-subjects design as in \cite{conati2021toward}. We recruited 39 participants by advertising at our university. Of these, 23 participants (9 females) were assigned to the experimental group where they used the personalized version of the ACSP explanation interface described in the previous section, and 16 participants (9 females) were assigned to the control group who used the original version of the explanation interface. 

Similar to \cite{conati2021toward}, participants were required to not be color-blind and to have basic knowledge of graph theory and algebra to learn concepts relevant to CSP.  In addition to these criteria, participants were required to have low N4C and low Conscientiousness. To find such LNLC participants, we asked those who replied to our advertisement to take standard online tests for N4C and Conscientiousness, namely the N4C test \cite{cacioppo1984efficient} (range +/- 36) and the TIPI personality test \cite{rothmann2003big} which includes scales for  Conscientiousness (range 1 to 7). These tests were also used in \cite{conati2021toward}. We selected participants who scored below the median for these two tests, which was computed using the N4C and Conscientiousness values from the 72 subjects in \cite{conati2021toward}, resulting in a median of 0 for N4C and 5.5 for Conscientiousness. We verified the reliability of our distributions by comparing them to those in the literature that use the same tests \cite{gosling2003very,thomson2016investigating}.

It should be noted that, initially, we aimed to replicate the size of the user distribution in \cite{conati2021toward} i.e., 30 users in the experimental condition and 17 in the control condition. However, our stricter selection criteria (requiring users to be LNLC) reduced the size of our target demographic by 70\%, which together with constraints related to the scheduling of the academic terms required us to settle for a slightly smaller sample size for the experimental group.
\begin{table}
    \centering
    \begin{tabular}{ll}
        \hline
        \multicolumn{2}{l}{\textbf{Items of Usefulness}} \\
        E1 & I would choose to have the explanations again in\\ & the future.\\
        E2 & I am satisfied with the explanations.\\
        E3 & The explanations were helpful for me.\\
         \multicolumn{2}{l}{\textbf{Items of Intrusiveness}} \\
        E4 & The explanations distracted me from my learning \\
        & task.\\
        E5 & The explanations were confusing.\\
        E6 & I found the explanations overwhelming.\\
        \hline
    \end{tabular}
    \caption{Perception of Explanations Questionnaire.}
    \label{tab:explanationquestion}
\end{table}
The study followed the experimental procedure in \cite{conati2021toward}, briefly summarized here. Users first studied a textbook chapter on the AC-3 algorithm, and then they wrote a pre-test on this material. Next, they watched a video on how to use the main functionality of the ACSP applet and then solved three CSPs using their respective versions of the ACSP applet (i.e., either with the personalized or the original explanation interface). After interacting with the ACSP applet, users took a post-test analogous to the pre-test. Finally, users were asked to complete two questionnaires. The first questionnaire elicited their subjective opinions on the usefulness and intrusiveness of the explanations (Table ~\ref{tab:explanationquestion})  to ascertain how the personalized version compared with the original one with respect to these subjective measures. Users in the control group who chose not to open the explanation interface did not fill out this questionnaire as they did not see the explanations. Instead, they were prompted to respond to an open-ended question: "Please describe why you did not access the explanations using the interface action: '\textit{Why was I delivered this hint?}'"

The second questionnaire elicited the users' subjective opinions on the usefulness, intrusiveness, and understanding/trust of the hints (Table ~\ref{tab:hintquestion}) to determine if the personalized explanations affect this perception. For both questionnaires, responses were provided on a Likert scale ranging from 1 ('Strongly Disagree') to 5 ('Strongly Agree'). 

The study took up to 2.5 hours and users were compensated with \$40. During the interaction with the ACSP, users’ gaze was tracked with a Tobii Pro X3 T-120 eye-tracker, which returns sequences of \textit{fixations}, namely, points of attention to the screen. Out of the 23 users in the experimental group, 3 did not receive any hints during their interaction with the ACSP applet and, thus, they are not involved in the analysis presented in the following sections, leaving us with 20 users in the experimental group moving forward. All users in the control group received at least one hint during the course of their interaction.

\begin{table}
    \centering
    \begin{tabular}{ll}
        \hline
        \multicolumn{2}{l}{\textbf{Items of Usefulness}} \\
        H1 & I would choose to have the hints again in the future.\\
        H2 & I am satisfied with the hints.\\
        H3 & The hints were helpful for me.\\
         \multicolumn{2}{l}{\textbf{Items of Intrusiveness}} \\
        H4 & The hints distracted me from my learning task.\\
        H5 & The hints were confusing.\\
        \multicolumn{2}{l}{\textbf{Items of Understanding/Trust.}} \\
        H6 & I understand why hints were delivered to me\\
        & in general.\\
        H7 & I understand why specific hints were delivered\\
        & me.\\
        H8 &  I trust the system to deliver appropriate hints. \\ 
        H9 & I agree with the hints that were delivered to me.\\
        H10 & The hints appeared at the right time. \\
        \hline
    \end{tabular}
    \caption{Perception of Hints Questionnaire.}
    \label{tab:hintquestion}
\end{table}

We evaluate the effectiveness of the personalized explanation interface for LNLC users by looking at (i) objective performance measures i.e., increase in user interaction with the explanation interface and learning from the pre-test to the post-test, and (ii) measures of subjective perceptions of explanations and hints. 
In our analyses, we use the Shapiro-Wilk test (SW) to test for data normality, a two-sample unpaired t-test to compare data that is normally distributed, and the Mann-Whitney U test otherwise. The Mann-Whitney U test is also used for ordinal Likert-scale data. For all the statistical tests, a critical value of 0.05 was used to determine significance. We remove outliers using U+/-2SD.  

\section{Results for objective measures}

\subsection{Effects on interaction with explanations} 
Conati et al., \shortcite{conati2021toward} evaluated user interaction with the explanation interface by analyzing the logged actions related to accessing explanation pages and the eye-tracking (ET) data to receive more information on a user’s attention to the explanations. Namely, they looked at the following measures:
\begin{itemize}
    \item Number of page types accessed: this measure shows to what extent users visit the 6 different types of pages available and it ranges from 0 to 6.
    \item Number of individual pages accessed: this measure accounts for the fact that each explanation page can be accessed multiple times.
    \item Total fixation duration (in seconds) on any explanation page accessed normalized by the total number of hints received: this measure captures the time  spent looking at the  explanations, normalized to account for the varied number of hints received by  different users. 
\end{itemize}

Because in our study the first explanation page is presented upfront to users in the experimental group, they do not need to perform an explicit action to access it. To account for this discrepancy, we analyze users' access to the first and remaining pages separately, and then we look at their attention measures.


In the control group, where the first page of the explanation interface needs to be explicitly opened, 7 out of 16 users (43.8\%) opened it. Of these 7, only 3 explored explanation pages beyond the first one. 

\begin{figure}[t]
 \centering
  \includegraphics[width=180px]{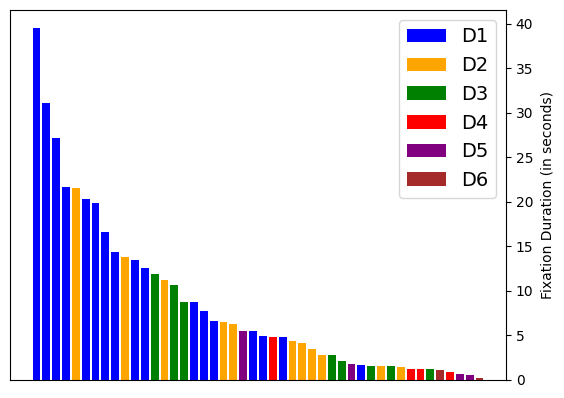}
  \caption{Distribution of fixation duration (in seconds) for all deliveries of the first explanation page to users in the experimental group. D'n' refers to the n$^{th}$ delivery of the page for a given user.}
  \label{fig:fix_dur_first}
\end{figure}

In the experimental group, for which the explanation interface appears upfront, all users looked at the first explanation page for 5 seconds or longer when they received the first hint, except for one user who glanced at it for 1.6 seconds. This information can be seen in  Fig. \ref{fig:fix_dur_first}, where the bars in blue ('D1') indicate the fixation duration on the first explanation page the first time it is delivered. The figure also shows the fixation durations on the page for subsequent deliveries.  These durations, not surprisingly, are higher for the first few occurrences (D1, D2, and D3) ranging from approximately 40s to 5s, and decrease for the subsequent occurrences (D4, D5, and D6) ranging from 5s to 0.8s. 
Out of the 20 users in the experimental group, 14 (70\%) accessed explanation pages beyond the first one, and they accessed an average of 5.1 pages (SD=2.9) and 2.7 page types (SD=1.1). Of the 3 users in the control group who accessed pages beyond the first one, one had slightly lower access than the average experimental user (2 pages and 2 page types accessed), one slightly higher (7 pages and 4 page types accessed), and one substantially higher (10 pages and 5 page types accessed). Interestingly, this last user showed negative learning gains. 

Evidently, the control group had a lower interaction with the explanation pages than the experimental group in terms of access actions. To ascertain how this lower interaction reflects on users' attention to the explanation pages, we first compare overall attention to the explanation interface (AE) as defined in \cite{conati2021toward}: the total fixation duration on any of the explanation pages divided by the total number of hints received.
For this analysis, we removed  3 users due to ET data loss (all from the experimental group) and 3 users for being outliers (one from the experimental and 2 from the control group),   leaving us with 16 users in the experimental group and 14 in the control. 

The AE data showed a significant departure from normality. Hence, we conducted a Mann-Whitney U test to compare the AE values of the experimental and control groups. The test revealed that the experimental group had a statistically higher AE (M=34.1s, SD=26.3s) than the control group (M=6.6s, SD=18.4s) with a large effect size (U (30) = 20, p=0.001, r\footnote{\label{u}Effect size range: $r$: 0.1 (small), 0.3 (medium), 0.5 (large)}=0.70).
This result indicates that, overall, users in the experimental group not only accessed more explanation pages but also paid more attention to them. 

Table \ref{tab:attention} reports the statistics for the AE values computed separately for the first explanation page and the following ones. The table shows comparable average AE values for the first page between the 6 control users who explicitly accessed it (13.0) and the 16 experimental users to whom it was delivered upfront (12.0). This suggests that showing the first explanation page upfront in the experimental condition elicits similar levels of attention to those shown by users who willingly chose to access the explanations in the control group.

The table also shows that the average AE to other pages is much higher for the 3 users in the control group than for the 14 users in the experimental group. However, this large difference is primarily driven by two control group users who have exceptionally large AEs, namely 171 seconds and 227 seconds respectively, where the latter AE is for the user who showed very high page accesses as discussed earlier. These two users are effectively outliers when AEs to other pages are computed over the experimental and control groups combined, and they both showed very low learning. 
In summary, the results in this section show that the personalized explanation interface succeeded in increasing LNLC users' interaction with it, in terms of both access and attention. 

\begin{table}
    \centering
    \begin{tabular}{lllll}
        \hline
        \textbf{Condition} &  \textbf{N} & \textbf{AE first page}& \textbf{N} & \textbf{AE other pages}\\
        & & \textbf{Mean (Sd)} &  & \textbf{Mean (Sd)}\\
        \hline
       Control & 6 & 13.0 (12.3) & 3 & 139.1 (108.0)\\
       Experimental & 16 & 12.0 (8.5) & 14 & 37.6 (53.0)\\
       
        \hline
    \end{tabular}
    \caption{Mean and standard deviation of AE for first and other pages.}
    \label{tab:attention}
\end{table}

To determine to what extent this increase is due to the confirmation box we added to warn users in the experimental group if they prematurely tried to close the explanation interface,  we analyzed users' responses to this box. Ten out of the 20 users in the experimental group received the confirmation box. Out of these 10, 7 dismissed it, while the remaining 3 stayed in the explanation interface for an additional 10, 20, and 28 seconds, and accessed 1, 3, and 5 additional pages respectively. These results indicate that, although the confirmation box was impactful for some users, it did not work for the majority who received it, suggesting that the upfront delivery of the explanation interface was the intervention that led to an increase in interaction with the explanations. 

\subsection{Effects on learning gains}
As in \cite{conati2021toward}, we measure users’ learning using percentage learning gains (PLG), namely, the difference between post-test and pre-test scores, divided by the absolute value of the maximum possible difference between the pre-test and post-test scores. 

\begin{figure}[t]
 \centering
  \includegraphics[width=150px]{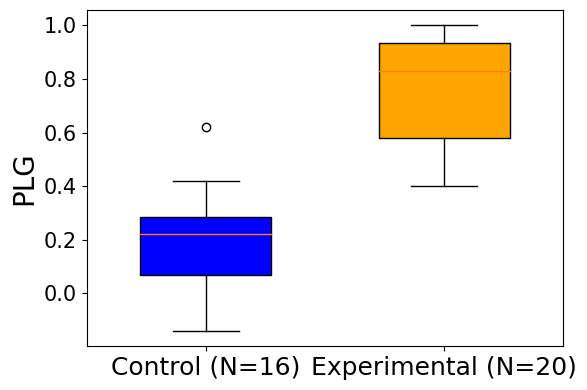}
  \caption{Effect of personalization on PLG.}
  \label{fig:plg}
\end{figure}

After removing one outlier from both control and experimental groups, the PLG values were found to be normally distributed. A two-sample unpaired t-test to compare the PLG values of the control and experimental groups revealed that the experimental group had a statistically higher PLG (M=0.7, SD=0.2) than the control group (M=0.2, SD=0.1) with a very large effect size (t (32) = -8.34, p$<$0.001, d\footnote{Effect size range:  Cohen’s d: 0.2 (small), 0.5 (medium), 0.8 (large), 1.2 (very large)}=2.88, as shown in Fig. ~\ref{fig:plg}). Together with the increased interaction of the experimental group with the explanation interface, this result indicates a possible relationship between users' amount of engagement with the explanations and their learning. To look more into this possible relationship, we performed a Spearman's rank correlation analysis of the AE and PLG values from all users (irrespective of whether they were from the control or experimental groups), as this correlation should not be affected by the delivery method. After removing 5 data points that were outliers (2 in PLG and 3 in AE), the analysis showed a significantly moderate correlation between AE and PLG ($r_s$(28)\footnote{Correlation range for Spearman's correlation: $r_s$: 0.4 to 0.59 ( moderate), 0.6 to 0.79 (strong), 0.8 to 1 (very strong)}=0.5, p=0.007).
Although correlation does not imply causation, this result suggests that it is worthwhile to further investigate whether higher interaction with the explanations (which is the focus of the personalized ACSP applet) leads to higher learning, e.g., by collecting more data to conduct a Structural Equation Modelling (SEM) analysis.

\section{Results on subjective measures} 

\begin{figure}[t]
 \centering
  \includegraphics[width=230px]{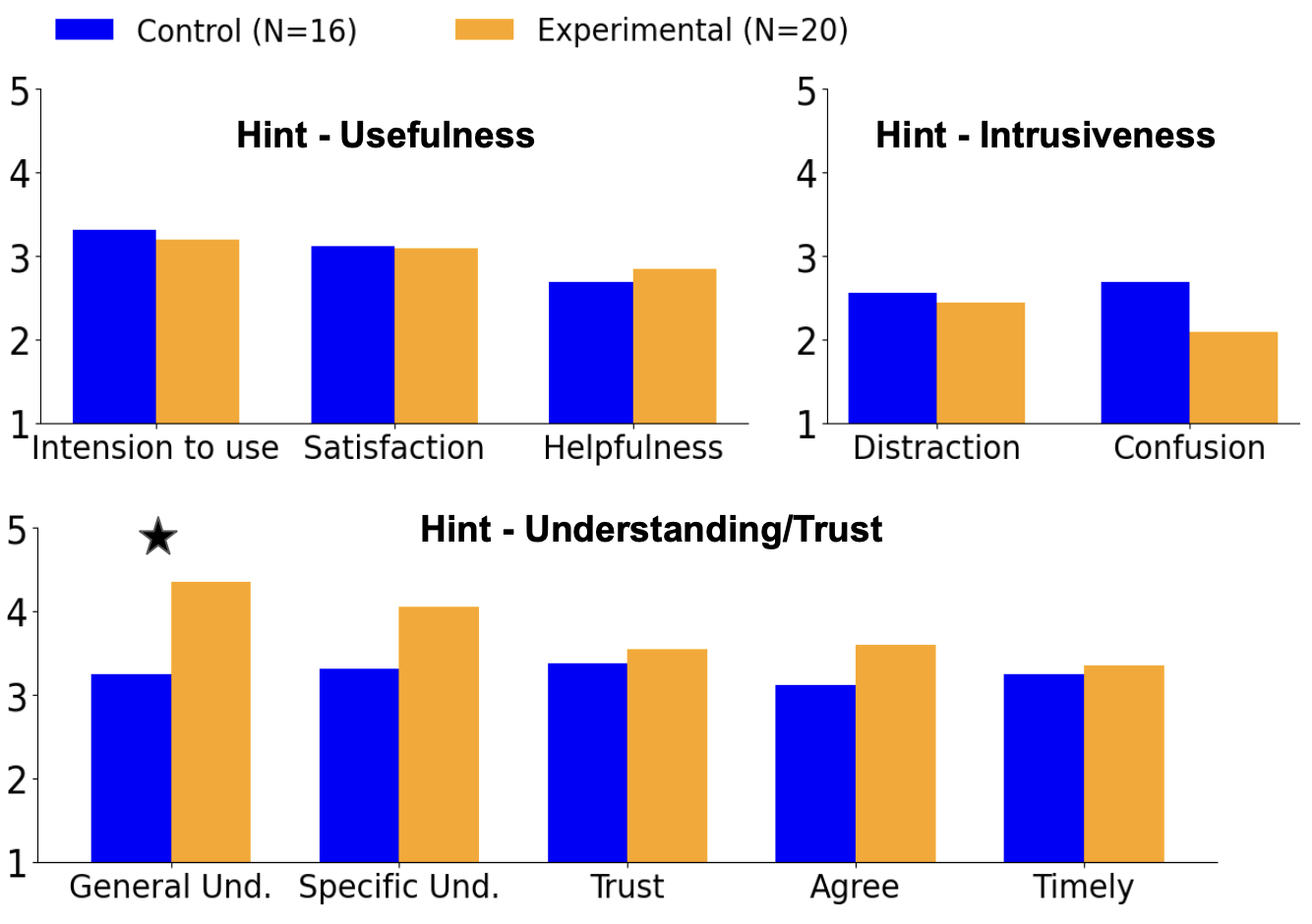}
  \caption{Subjective ratings of hints. "*" indicates a statistical difference between the control and experimental  groups.}
  \label{fig:hintratings}
\end{figure}

\textbf{Subjective ratings of hints}:
To evaluate the impact of personalizing the explanation interface on users' perception of the ACSP hints, we compared subjective ratings from the questionnaire on hint perception (see Table \ref{tab:hintquestion}) between the experimental and control groups.
These ratings are shown in Fig. \ref{fig:hintratings}, where they are grouped into three sets of questionnaire items: items related to Usefulness, Intrusiveness, and Understanding/Trust. All comparisons are done with a Wilcoxon-Mann-Whitney test, which is suitable for ordinal ratings. We observe that the ratings for Usefulness and  Intrusiveness are similar for both groups. For Understanding/Trust, we find a significant difference with a large effect size for General Understanding (U=238.5, p=0.04, r$^1$=0.74 ) and a non-significant trend with a large effect size for Specific Understanding (U=222, p=0.09, r$^1$=0.69), both with higher values for the experimental group.

 
\begin{figure}[t]
 \centering
  \includegraphics[width=255px]{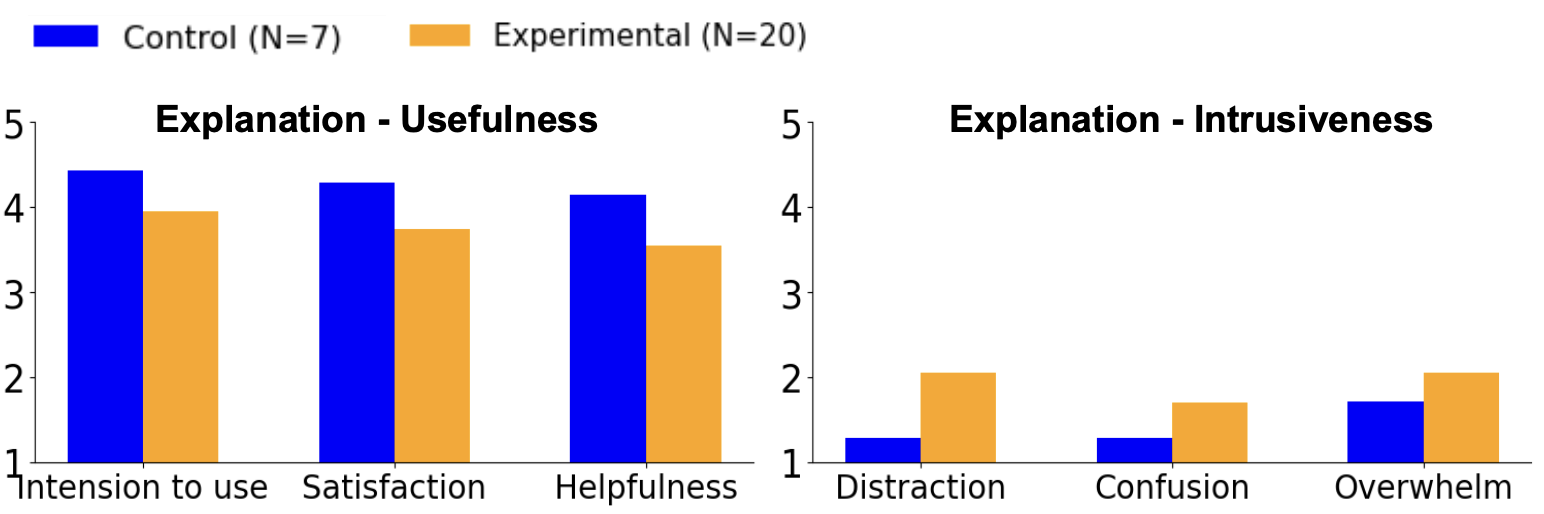}
  \caption{Subjective ratings of explanations.}
  \label{fig:explratings}
\end{figure}

\textbf{Subjective ratings of explanations}:
Fig. \ref{fig:explratings} shows the average ratings provided by users in the two groups for the questionnaire (Table \ref{tab:explanationquestion}) probing users' perception of the explanation interface's usefulness and intrusiveness. The figure illustrates a trend wherein the control group tends to receive better ratings (higher for usefulness and lower ratings for intrusiveness) compared to the experimental group. None of these trends are statistically significant, but there are large effect sizes for confusion (r$^1$=0.62) and distraction (r$^1$=0.69), which is not surprising because the explanation interface takes up screen space and covers parts of the CSP question being solved. Regardless,  the average ratings provided by the experimental group still translate to positive sentiments for all aspects i.e., greater than 3.5 for ratings on usefulness and lower than 2 for intrusiveness.

\section{Discussion}

Our findings show that personalizing the explanation interface  for AI-driven hints in the ACSP applet for LNLC users significantly increases their interaction with the explanations. Specifically,  only about 44\%  of the 16 control users accessed the explanation interface and only 3 went beyond the first explanation page. In contrast, 70\% of the 20 users in the experimental group willingly accessed pages beyond the one presented upfront.  Our results also show that users in the experimental group not only accessed more explanation pages but also paid more attention to them, resulting in higher overall attention to explanation pages compared to the control group.   
  
During the study, we prompted the  9 users in the control group who did not engage with the explanation interface about their reasons and received the following responses: 2 wished to concentrate on solving the CSP problem without being distracted, 3 did not notice the button to open the explanation interface, and 4 did not perceive a need for explanations. The latter two issues could possibly be addressed by the personalized explanation interface that was used by the experimental group where the first explanation page is delivered upfront with the hint, which eliminates the need to click on a button to open the explanation interface and potentially demonstrates the relevance of the explanations to those who don't perceive it a priori.

Our findings also revealed that the users for whom the explanations were personalized showed significantly higher learning gains. This finding is of paramount importance as the principal objective of an ITS is to improve learning. 

Finally, we found that the personalized explanation interface improves users' perceived understanding of the AI-driven hints compared to the original explanation interface. This increased perceived understanding is consistent with the fact that these users interacted with the explanations more and the explanations justify the reasoning underlying the hints.  

However, despite the increase in learning gains and perceived hint understanding, users in the experimental group rated the explanations relatively worse than those in the control group, especially  for   distraction and confusion. At the end of the study, we asked the 6 users in the experimental group who gave ratings higher than 3 for these items the reasons for the poor ratings. We found that: 2 users considered the explanations to be too verbose, 2 deemed the hint to be insignificant and, thus, perceived the corresponding explanations to be not meaningful, and 2 found that the hints and explanation interface appeared too often. The first two reasons are inherent to the content of the explanation and hint, which was unchanged between the personalized and the original explanation interface. The third concern regarding the frequency of explanations is particularly relevant in the personalized interface, where the frequent delivery of bulky explanations along with the hints can be easily perceived as intrusive. One possible way to address this concern is to monitor users' reactions to the upfront delivery (e.g., attention to the first explanation page)  and make the explanations accessible on demand if their reactions are negative (e.g., decreasing interest).


 In addition to delivering the first page of the explanation interface upfront, we also added a confirmation box to the personalized ACSP applet to increase the time spent by users who prematurely tried to close the explanation interface. The confirmation box was found not to be very impactful. Upon further investigating the possible reasons for this result, we found that 4 of the 7 users who dismissed the confirmation box had previously accessed the explanation interface for longer than 27 seconds and only received the confirmation box during their subsequent accesses.  As these users had already seen the explanations and parts of the explanation content stay the same for different hints, they likely needed less time to process subsequent explanations. Hence, they should not have been asked to stay longer. To address this issue, the threshold for triggering the confirmation box should be adjusted to take into account previous viewing of the explanations.   

\section{Conclusion and future work}
In this work, we investigated personalizing the explanations provided by the ACSP tutoring system to justify the hints that it generates to foster student learning. The personalization targets students with low levels of Need for Cognition and Conscientiousness, and aims to enhance their engagement with the explanations based on prior findings that these students do not naturally engage with the explanations but they would benefit from them if they do. The personalization consists of delivering a first level of explanation together with the corresponding hint and using prompts  (a confirmation box) to warn students when they try to leave the explanations too quickly. We  showed that  the personalized explanations significantly increased users’ interaction with the explanations, their perceived understanding of the hints, and their learning. However, the study also revealed that the prompts to stay  had limited impact on these results. 

Our work contributes new evidence to the  benefits of Personalised XAI, showing that personalized explanations can help users with specific traits. Previous work demonstrated   these benefits for relatively simple explanations in a less cognitively demanding task (interacting with a musing recommender)  or, like us, in the more complex task of learning  but still involving very simple one-shot explanations.

Moving forward, in the short term we will investigate the  improvements suggested by our study to reduce the possible distraction/confusion of the upfront delivery and to increase the usefulness of the confirmation box. In the long term, we will investigate personalizing the ACSP explanations to a user's level of Reading Proficiency, another user characteristic identified as relevant by previous work. Moreover, we want to dynamically personalize the ACSP explanations  by building classifiers that can detect relevant user traits during interaction, based on existing work on the real-time detection of personality traits and cognitive abilities from multi-modal data such as interface actions (e.g. \cite{kuster2018predicting}) and eye-tracking data (e.g.\cite{millecamp2020s}).

\bibliographystyle{named}
\bibliography{ijcai24}

\end{document}